\documentclass[conference]{IEEEtran}
\IEEEoverridecommandlockouts
\usepackage{comment}
\usepackage{cite}
\usepackage{amsmath,amssymb,amsfonts}
\usepackage{algorithmic}
\usepackage{graphicx}
\usepackage{subcaption}
\usepackage{textcomp}
\usepackage{xcolor}
\usepackage[ruled]{algorithm2e}
\def\BibTeX{{\rm B\kern-.05em{\sc i\kern-.025em b}\kern-.08em
    T\kern-.1667em\lower.7ex\hbox{E}\kern-.125emX}}
\begin{document}

\title{Harmonic Networks with Limited Training Samples \\
\thanks{This work is supported by the ADAPT Centre for Digital Content Technology funded under the Science Foundation Ireland Research Centres Programme grant 13/RC/2106 and co-funded under the European Regional Development Fund. The second author is also supported by the European Union’s Horizon 2020 research and innovation programme under the Marie Sklodowska-Curie grant agreement No.713567.}
}

\author{
\IEEEauthorblockN{Matej Ulicny, \qquad\qquad Vladimir A. Krylov, \qquad\qquad Rozenn Dahyot}
\IEEEauthorblockA{ADAPT Centre, School of Computer Science \& Statistics, Trinity College Dublin, Dublin, Ireland\\
\{ulinm, Vladimir.Krylov, Rozenn.Dahyot\}@tcd.ie}
}

\maketitle

\begin{abstract}
Convolutional neural networks (CNNs) are very popular nowadays for image processing. CNNs allow one to learn optimal filters in a (mostly) supervised machine learning context. However this typically requires abundant labelled training data to estimate the filter parameters. Alternative strategies have been deployed for reducing the number of parameters and / or filters to be learned and thus decrease overfitting. In the context of reverting to preset filters, we propose here a computationally efficient harmonic block that uses Discrete Cosine Transform (DCT) filters in CNNs. In this work we examine the performance of harmonic networks in limited training data scenario. We validate experimentally that its performance compares well against scattering networks that use wavelets as preset filters.
\end{abstract}


\section{Introduction} 
\label{sec:intro}

We have recently proposed a new form of neural network layer called harmonic block \cite{Ulicny18} that relies on using windowed cosine transform at several frequencies in lieu of learned filters. This harmonic block only involves learning weights for combining several frequency responses together in the frequency domain. Furthermore, uninformative frequencies can be dropped out to improve the computational complexity of the network without compromising performance, i.e. compression~\cite{Ulicny18}. 
This paper extends further the proposed harmonic block by: 1) showing how it relates to the modified discrete cosine transform when considering overlap in computing convolution, 2) proposing an improved, computationally more efficient implementation, and 3) showing that the CNNs using the harmonic block outperform scattering network, based on the use of wavelet-based filters~\cite{Bruna13,Oyallon18} when training data is scarce. The PyTorch implementation of the harmonic block is provided at \textit{\url{https://github.com/matej-ulicny/harmonic-networks}}.

The rest of the paper is organised as follows. We first review the related literature (Sec.~\ref{sec:rel_work}) and present the harmonic block (Sec.~\ref{sec:trans}). We then report the experimental validation (Sec.~\ref{sec:exp}) and conclusions of the study (Sec.~V). 

\section{Related work}
\label{sec:rel_work}

\subsection{DCT \& CNNs}

Wang and Zhang~\cite{Wang16} propose a double JPEG
compression detection algorithm based on a convolutional neural network (CNN) to detect tampered area for  image forensics. The 1-dimensional CNN is designed to classify histograms of discrete cosine transform (DCT) coefficients, which differ between single-compressed areas (tampered areas) and double-compressed areas (untampered areas) ~\cite{Wang16,Amerini17,Barni17}. Alternatively, raw DCT (discrete cosine transform) coefficients from JPEG images has also been proposed as input of a  2-dimensional CNN ~\cite{Li17}.
Spectral image representations combined with neural networks have also been used for object recognition. For instance, truncation of DCT coefficients has been shown to speed up training of fully connected sparse autoencoders~\cite{Zou14} and improve face recognition with linear discriminant analysis and radial basis function network~\cite{Er05}. DCT transform has been used in conjunction with CNNs for image classification as an input pre-processing step~\cite{Ulicny17,Gueguen18}. Ghosh and Chellappa~\cite{Ghosh16} transformed feature maps inside the CNN pipeline and noted convergence improvements. 

\subsection{Wavelets \& CNNs}

Common approach in literature is to use wavelet transform to extract invariant features prior to classification. One such example is the Scattering convolution network composed of complex Morlet wavelet filters~\cite{Bruna13} and a PCA or SVM classifier. Wavelet responses were also used with NN-based classifier~\cite{Said16}, or with a set of CNNs each operating on exclusive frequency sub-band~\cite{Williams16}. Silva et al. used wavelet filters to enhance edges prior to CNN processing~\cite{Silva18}.
Rotation and scale invariant wavelet based scattering networks with subsequent CNN were formulated in~\cite{Oyallon18,Singh17}. These hybrid networks were shown to reach comparable classification accuracy to deeper CNNs.  

Several studies incorporated wavelets in CNN computational graphs. New feature pooling strategies were designed based on fast Fourier transform~\cite{Rippel15} or fast wavelet transform~\cite{Williams18}. Haar wavelet responses of the input image have been concatenated to features at different stages of CNN to address texture classification~\cite{Fujieda17}. Lu et al.~\cite{Lu18} designed a similar approach for medical image segmentation, however based on dual-tree complex wavelets. Robustness to scale and orientation of CNN is increased by modulating learned filters by a set of Gabor filters~\cite{Luan18}. Rotation equivariance of learned features was accomplished by incorporated complex circular harmonics into CNNs~\cite{Worrall17}. Jacobsen et al. proposed to learn convolution filters as a composition of Gaussian derivative filter basis~\cite{Jacobsen16}.

\subsection{Compressing CNNs}

Compression of neural networks has received a lot of attention from researchers. Jaderberg et al.~\cite{Jaderberg14} approximated full-rank CNN filters by separable rank-1 filters. DCT transform has been used for model compression, to cluster weights into buckets based on their DCT representation~\cite{Chen16}, or to represent weights as residuals from their cluster centers in DCT domain~\cite{Wang16b}.


\section{Harmonic block}
\label{sec:trans}

\subsection{Overlapping cosine transform} \label{sec:overlap_dct}

DCT computed on overlapping windows is also known as Lapped Transform or Modified DCT (MDCT), equivalent to our harmonic block using strides. The overlapped DCT has a long history in signal compression and reduces artefacts at window edges~\cite{Tran03}. Dedicated strategies for efficient computations have been proposed~\cite{Tran03}, including algorithms and hardware optimisations. Our current implementation uses standard deep learning libraries (PyTorch) and is not currently taking full advantage of these more  advanced  DCT implementations. 

DCT transform is equivalent to the discrete Fourier transform of real valued functions with even symmetry within twice larger window. DCT lacks imaginary component given by the sine transform of real valued odd functions. However, harmonic block allows convolution with DCT basis with arbitrary stride creating redundancy in the representation. Ignoring the boundary limitations, sine filter basis can be devised by shifting the cosine filters. Given the equivariant properties of convolution, instead of shifting the filters the same result is achieved by applying original filters to the shifted input. Considering DCT-II formulation:
\begin{equation} \label{eq:dct}
F_{k} = \sum_{n=0}^{N-1} x_n \cos{\left[\frac{\pi}{N} \left(n+\frac{1}{2}\right)k\right]}
\end{equation}
a corresponding sine transform is
\begin{equation} \label{eq:dst}
G_{k} = \sum_{n=0}^{N-1} x_n \sin{\left[\frac{\pi}{N} \left(n+\frac{1}{2}\right)k\right]}
\end{equation}
which is equivalent to
\begin{equation} \label{eq:dst_to_dct}
G_{k} = \sum_{n=0}^{N-1} x_n \cos{\left[\frac{\pi}{2}+2\pi z-\frac{\pi}{N} \left(n+\frac{1}{2}\right)k\right]}.
\end{equation}
The shift given by $\pi/2 + 2\pi z$ for any $z \in \mathbb{Z}$ can be directly converted to shift in pixels applied to data $x$. After simplification, sine transform can be expressed as \begin{equation} \label{eq:dct_shifted}
G_{k} = \sum_{n=0}^N{x_n \cos{\left[ \frac{\pi}{N} \left( n-\frac{N(1+4z)}{2k}+\frac{1}{2} \right) k \right]}}
\end{equation}
which is equivalent to the cosine transform of the image shifted by $\delta=N\left(1+4z\right)/2k$ defined in~\eqref{eq:dct_shifted_data}. 
\begin{equation} \label{eq:dct_shifted_data}
F_k[\delta]=\sum_{n=0}^N{x_{n+\frac{N(1+4z)}{2k}} \cos{\left[ \frac{\pi}{N} \left( n+\frac{1}{2} \right) k \right]}}.
\end{equation}
This value represents the stride to shift the cosine filters to capture correlation with sine function.

\subsection{Definition of harmonic block} \label{sec:trans/harm_block}
The harmonic block~\cite{Ulicny18} is designed to replace fully learned convolution of multidimensional input features $h^{l-1}$. Input channels $h_n^{l-1}, n \in \left\{ 0..N-1 \right\}$ are convolved using the DCT basis functions $\phi_{u,v}$ given size of the desired receptive field $K \times K$:
\begin{equation} \label{eq:dct_filt}
\phi_{u,v}\left(x,y\right) = \cos{\left[\frac{\pi}{K} \left(x+\frac{1}{2}\right)u\right]} \cos{\left[\frac{\pi}{K} \left(y+\frac{1}{2}\right)v\right]}.
\end{equation}
Specifically, we employ $L1$-normalised filters $\psi_{u,v}\in \mathcal{R}^{K \times K}$:
\begin{equation}
    \psi_{u,v} = \frac{\phi_{u,v}}{\left\lVert \phi_{u,v} \right\rVert_1}.
\end{equation}
Due to properties of natural images, high frequency responses are generally of lower magnitude. Employing batch normalization (BN) on DCT coefficients of the RGB channels has been found useful~\cite{Ulicny18} for propagating energy of the whole spatial-frequency spectrum. Output features $h_m^l, m \in \left\{ 0..M-1 \right\}$ are learned as superpositions of the DCT coefficients, described in detail in Algorithm~\ref{alg:harm_block}, where the learned parameters inside each harmonic block are denoted as $w \in \mathcal{R}^{M \times N \times K \times K}$.

\begin{algorithm}[!b]
 \KwIn{$h^{l-1}$}
 \For{$n \in \left\{ 0,\cdots,N-1 \right\}$}{
  $z^l_{n,u,v} \leftarrow \sum_{u=0}^{K-1}{\sum_{v=0}^{K-1}{\psi_{u,v}* *\, h^{l-1}_n}}$\\ 
  \If{normalize}{
   $\mu^l_{n,u,v},{\sigma^l_{n,u,v}} \leftarrow$ estimate mean and standard deviation of $z^l_{n,u,v}$ over the batch dimension
   $z^l_{n,u,v} \leftarrow \frac{\left(z^l_{n,u,v} - \mu^l_{n,u,v}\right)} {\sigma^l_{n,u,v}}$
  }
 }
 \For{$m \in \left\{ 0 \dots M-1 \right\}$}{
  $h^l_{m} \leftarrow \sum_{n=0}^{N-1}{\sum_{v=0}^{K-1}{\sum_{u=0}^{K-1}{w_{m,n,u,v}\ z_{n,u,v}}}}$
 }
\KwOut{$h^l$}
\caption{Harmonic block}\label{alg:harm_block}
\end{algorithm}

The downside of Algorithm~\ref{alg:harm_block} is that in order to be executed in parallel, extra memory has to be allocated to store the responses of DCT filters at every layer. Since most of the blocks do not need to use BN they become linear. Hence DCT transform and linear combination can be merged into a single linear operation. In other words, equivalent features can be obtained by factorizing filters as linear combination of DCT basis functions. Therefore we propose here Algorithm~\ref{alg:mem_eff_harm_block} that is a more efficient alternative to Algorithm~\ref{alg:harm_block}. This reformulation is similar to structured receptive field~\cite{Jacobsen16} utilizing different basis functions. The theoretical number of multiply-add operations compared to the standard convolutional layer increases by a factor of $K^2/M$ for Algorithm 1, and by $K^2/AB$ for Algorithm 2, where the input image size for the block is $A\times B$. The experimental performance of the two algorithms is compared in Section~\ref{sec:exp/req}.

\begin{algorithm}[!h]
 \KwIn{$h^{l-1}$}
 {Define updates} $g \in \mathcal{R}^{M \times N \times K \times K}$\;
 \For{$m \in \left\{ 0..M-1 \right\}$}{
  \For{$n \in \left\{ 0..N-1 \right\}$}{
   $g^l_{m,n} \leftarrow \sum_{u=0}^{K-1}{\sum_{v=0}^{K-1}{w_{m,n,u,v}\ \psi_{u,v}}}$\;
  }
 }
 $h^l \leftarrow g^l**h^{l-1}$\;
 \KwOut{$h^l$}
 \caption{Memory efficient harmonic block} \label{alg:mem_eff_harm_block}
\end{algorithm}

\begin{figure}[b]
\begin{center}
   \includegraphics[width=0.6\linewidth]{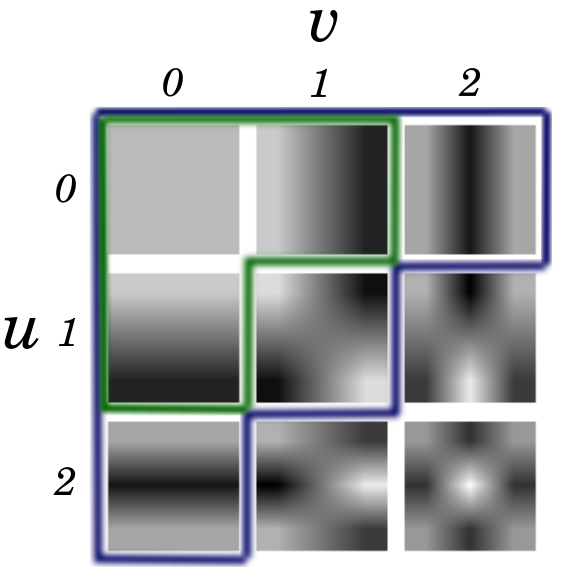}
\end{center}
   \caption{DCT filter bank employed in harmonic blocks. Blue (green) color filters are used to produce features for $\lambda$=3 ($\lambda$=2).}
\label{fig:filters}
\end{figure}

Control over the filters allows one to achieve reduced computational complexity by selecting subsets of filters to approximate the signal. A $\lambda$-subset is a collection of all filters $\psi_{u,v}$ such that their indices $u,v$ satisfy the condition $u+v<\lambda$. Fig.~\ref{fig:filters} shows example of some subsets of 3-by-3 DCT filters.

\section{Experimental evaluation} \label{sec:exp}

\subsection{Computational requirements} 
\label{sec:exp/req}
Firstly we compare the two implementations of a harmonic block, see Sec.~\ref{sec:trans/harm_block}. Experiment is conducted on well performing wide residual network (WRN)~\cite{Zagoruyko16} trained on CIFAR10 dataset. The baseline WRN 16-8 (for architecture details and training procedure see~\cite{Zagoruyko16}) with dropout rate 0.2 is compared with harmonic WRN with all convolution layers replaced by blocks defined in Algorithm~\ref{alg:harm_block} with additional BN in the first block. 
The network runtime and memory requirements for Algorithm~\ref{alg:harm_block} far exceed those of the baseline WRN (implemented via deep learning framework and run on GPUs)
despite being more flexible and having similar amount of arithmetic operations, see discussion in~\cite{Ulicny18}. Fully harmonic WRN based on Algorithm~\ref{alg:mem_eff_harm_block} (except the first layer due to the presence of BN) largely outperforms Algorithm~\ref{alg:harm_block} and shows only a modest increase in runtime and memory usage over the baseline WRN~\cite{Zagoruyko16} while having competitive performance.

\begin{table}[!t]
\centering
\caption{Computational requirements of harmonic block implementations on CIFAR10. Accuracy shown is an average over 5 runs with empirical one standard deviation interval.} \label{tab:alg_req}
 \begin{tabular}{|l|c|ccc|}
  \hline
  Model & Ref. & GPU  & epoch  & acc. \\
 &&  mem. & runtime &  \\
  \hline
  WRN 16-8 &~\cite{Zagoruyko16} & 2.8GB & 45.0s & 95.61$\pm$0.14 \\
  Harm WRN 16-8 (Alg.~\ref{alg:harm_block}) &~\cite{Ulicny18} & 6.6GB & 123.4s & 95.56$\pm$0.04 \\
  Harm WRN 16-8 (Alg.~\ref{alg:mem_eff_harm_block}) &  & 2.9GB & 56.8s & 95.62$\pm$0.09 \\
  \hline
 \end{tabular}
\end{table}

\subsection{Overlapping DCT experiments} \label{sec:exp/overlap}

\begin{figure*}
 \centering
 \begin{subfigure}[b]{0.323\linewidth}
  \includegraphics[width=\textwidth]{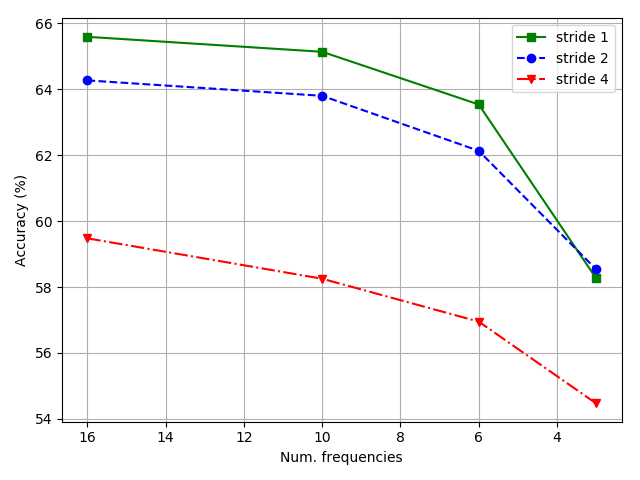}
  \caption{4x4 replicated features} \label{fig:stride_repl}
 \end{subfigure}
 \begin{subfigure}[b]{0.323\linewidth}
  \includegraphics[width=\textwidth]{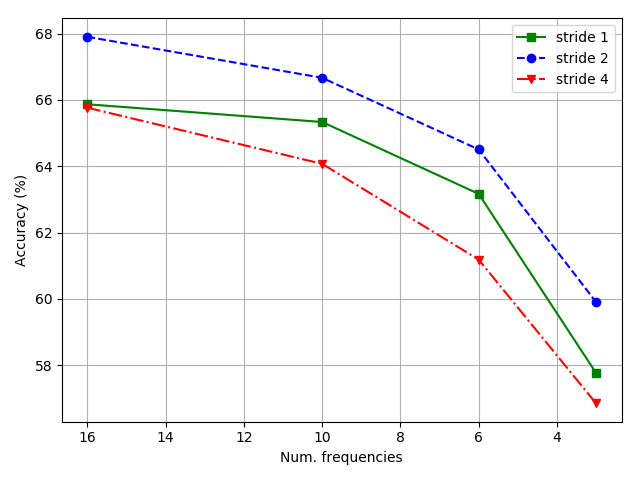}
  \caption{4x4 balanced block} \label{fig:stride_bal4}
 \end{subfigure}
  \begin{subfigure}[b]{0.323\linewidth}
  \includegraphics[width=\textwidth]{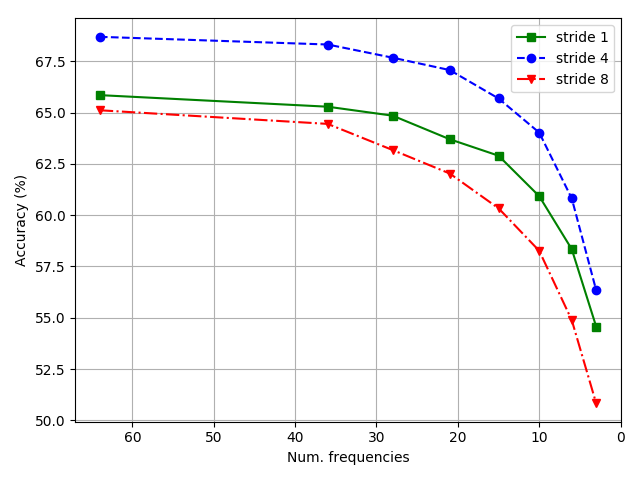}
  \caption{8x8 balanced block} \label{fig:stride_bal8}
 \end{subfigure}
   \caption{Accuracy degradation of models with different strides when truncating number of DCT coefficients. Stride 1 (green), half window stride (blue) and full window stride (red) are compared. Reported values are averaged over 5 runs.} \label{fig:stride}
\end{figure*}

In Section~\ref{sec:overlap_dct} we  demonstrated that the discrete sine transform can be inferred from the DCT on overlapping blocks. Here we show experimentally the benefits of DCT transform with overlapping windows by using overcomplete representation with strides of 1 pixel or fixing stride to the half of the window size.
Effect of striding is evaluated on a shallow harmonic network composed of only one normalized harmonic block with 4x4 receptive field, followed by a Rectified Linear Unit (ReLU) activation and connected to a fully connected layer with softmax classifier. This simple architecture allows one to clearly see the contribution of striding. The network is trained with SGD using learning rate 0.01, Nesterov momentum 0.9, weight decay 0.0005 and batch size 128 for 30 epochs decaying learning rate by factor 10 halfway.
Since striding reduces the spatial resolution of the features, to match the model complexity, lower dimensional features are resized to have size of features produced by stride 1. As expected, network without overlapping windows performs notably worse even with full spectrum (see Fig.~\ref{fig:stride_repl}). 

In order to compare models with similar numbers of parameters, instead of replicating features, networks with larger stride employ a higher number of output features: 200 for non-overlapping, 50 for half-window overlap in contrast to 16 when using stride 1. The same experiment is performed using 8x8 filters learning 625, 200 and 16 feature maps respectively. In this setting network with stride 1 and with full window stride perform comparably on full spectrum as can be seen on Fig.~\ref{fig:stride_bal4} and Fig.~\ref{fig:stride_bal8}, but performance degrades more rapidly for non-overlapping filters as the visual spectrum shrinks. The best result was obtained when using half window stride.

\subsection{Limited Data} \label{sec:exp/lim_data}

Deep neural networks require abundant data to achieve high accuracy. It has been shown in~\cite{Bruna13,Oyallon18} that  scattering network using geometric priors can learn better discrimination boundaries when presented with a small subset of training samples. We demonstrate capabilities of harmonic networks when learning from limited subsets of data on three datasets.

\subsubsection{MNIST} \label{sec:lim_data/mnist}
Bruna and Mallat~\cite{Bruna13}  have chosen a dataset of handwritten digits to test their fully handcrafted scattering network with respect to stability to deformations and classification performance on data subsets. We compare our harmonic network to the ``classical'' CNN, learned depth-separable convolution network and to the fully handcrafted scattering network~(as reported by~\cite{Bruna13}).
Table~\ref{tab:lim_data_mnist} shows the harmonic network achieves the lower classification error for all sizes of the training set. The baseline network is composed of 3 convolution layers with 32, 64 and 128 $3 \times 3$ filters, respectively, and with overlapping average pooling between them. Convolutional layers are followed by a fully connected layer with 512 neurons. Batch normalization and ReLU are applied after each layer. The harmonic network uses the same configuration replacing convolution with harmonic block while using additional BN in the first block. Harmonic networks are also compared to the depth-separable convolution network that has the same structure but has randomly initialized learnable filters instead of DCT filters. Training is done with SGD for 30 epochs with learning rate 0.1 reduced after every 10 epochs by a factor 10. Weight decay ranges from 0.0005 (for training size 300) to 0.05 (training size 60000). Harmonic networks outperform other networks in all configurations, see Tab.~\ref{tab:lim_data_mnist}.  

\begin{table}
\centering
\tabcolsep = 1.1mm
\caption{Classification errors in \% (median of 21 runs) on subsets of MNIST dataset for harmonic network and benchmarks.} \label{tab:lim_data_mnist}
 \begin{tabular}{|c|cccc|}
  \hline
  \footnotesize{Training size} & \footnotesize{Scat. net.~\cite{Bruna13}} & \footnotesize{Conv. net.} & \footnotesize{Sep. conv. net.} & \footnotesize{Harm. net.} \\
  \hline
  300 & 4.7 & 3.9 & 4.67 & \textbf{3.71} \\
  1000 & 2.3 & 1.88 & 1.91 & \textbf{1.84} \\
  2000 & 1.3 & 1.39 & 1.35 & \textbf{1.21} \\
  5000 & 1.03 & 0.97 & 1.06 & \textbf{0.86} \\
  10000 & 0.88 & 0.7 & 0.76 & \textbf{0.65} \\
  20000 & 0.58 & 0.59 & \textbf{0.57} & \textbf{0.57} \\
  40000 & 0.53 & 0.48 & 0.47 & \textbf{0.45} \\
  60000 & 0.43 & 0.44 & 0.46 & \textbf{0.38} \\
  \hline
 \end{tabular}
\end{table}

\begin{table}
\centering
\caption{Average classification accuracy $\pm$ standard deviation of 5 runs on subsets of CIFAR10.} \label{tab:lim_data_cifar}
\begin{tabular}{|l|c|c|c|c|}
 \hline
 \textbf{Method} & \textbf{100} & \textbf{500} & \textbf{1000} & \textbf{Full} \\
 \hline
 WRN 16-8 & 34.4$\pm$1.8 & 52.2$\pm$1.8 & 62.8$\pm$0.7 & \textbf{95.6} \\
 Scat + WRN~\cite{Oyallon18} & \textbf{38.9$\pm$1.2} & 54.7$\pm$0.6 & 62.0$\pm$1.1 & 93.1 \\
 Harm WRN 16-8 & 37.7$\pm$1.9 & 58.2$\pm$1.4 & 67.0$\pm$0.4 & \textbf{95.6} \\
 Harm WRN 16-8 $\lambda=3$ & 37.9$\pm$2.4 & \textbf{58.4$\pm$0.9} & \textbf{67.2$\pm$0.5} & \textbf{95.6} \\
 Harm WRN 16-8 $\lambda=2$ & 37.2$\pm$1.7 & 57.0$\pm$1.0 & 65.9$\pm$0.8 & 95.3 \\
 \hline
\end{tabular}
\end{table}

\subsubsection{CIFAR10}

We replicate the experiment in~\cite{Oyallon18} and train harmonic network on random subsets of CIFAR10 dataset with size 100, 500 and 1000 samples preserving equal number of labels per class. Harmonic WRN 16-8 with dropout rate 0.2 is trained as in~\cite{Oyallon18}. Harmonic layers relying on combinations of fixed filters give advantage on limited data compared to fully learned CNNs and to scattering CNN hybrids\footnote{The exact subsets used to train scattering CNN hybrids are not known, we report the numerical results from~\cite{Oyallon18}.} except for the smallest training dataset, see Tab.~\ref{tab:lim_data_cifar}. 

\subsubsection{STL10}

\begin{table}
\centering
\caption{Average classification accuracy $\pm$ standard deviation of 5 runs on STL10 (batch size 32).} \label{tab:stl}
\begin{tabular}{|l|c|c|c|}
 \hline
 \textbf{Method} & \textbf{10-folds} & \textbf{all} \\
 \hline
 WRN 16-8 & 73.50 $\pm$ 0.87 & 87.29 $\pm$ 0.21 \\
 Scat + WRN~\cite{Oyallon18} & 76.00 $\pm$ 0.60 & \hspace{-1.04cm} 87.60  \\
 Harm WRN 16-8 & 76.95 $\pm$ 0.93 & \textbf{90.45 $\pm$ 0.12} \\
 Harm WRN 16-8 $\lambda=3$ & 76.65 $\pm$ 0.90 & 90.39 $\pm$ 0.08 \\
 Harm WRN 16-8 progressive $\lambda$ & \textbf{77.19 $\pm$ 1.02} & 90.28 $\pm$ 0.20 \\
 \hline
\end{tabular}
\end{table}

STL10~\cite{STL10} is a natural image dataset similar to CIFAR10. Images are 96$\times$96 and only 5000 training images are labeled. The large set of provided unlabeled images is not utilized in this experiment. We design harmonic WRN 16-8 model (based on Algorithm~\ref{alg:mem_eff_harm_block}) for this task with several necessary modifications. The first layer uses stride 2, and the feature resolution at the final stage is 12$\times$12. We apply dropout 0.3 inside residual blocks and train the network on the whole training set with learning rate of 0.1 decayed by factor 0.2 after 300, 400, 600, 800 epochs, and stopping the training after 1000. The baseline network design and training procedure is similar to~\cite{Devries17} that uses additional cutout regularization and reports 87.26\% $\pm$ 0.23 on test set containing 8000 images when trained on batches of 128 images. The harmonic WRN 16-8 achieves 88.1\% $\pm$ 0.23 trained with the same settings. Decreasing the batch size to 32 improves our result to 90.45\% surpassing the deeper scattering WRN~\cite{Oyallon18} by nearly 3\%. Furthermore, when only predefined folds of 1000 samples serve as the training data, we obtain the best accuracy by progressively reducing the number of used frequencies along with spatial resolution: full filter bank is applied on features of size 48$\times$48, filters with $\lambda=3$ on 24$\times$24 and finally $\lambda=2$ if features are 12$\times$12. The results of STL10 experiments are summarised in Tab.~\ref{tab:stl}.

\section{Conclusion}

We have proposed a computationally efficient alternative to the original harmonic block based on DCT~\cite{Ulicny18}. The implementation is characterized by a very small increase in the number of multiply-add operations compared to a standard convolutional layer, thus enabling the wider use of harmonic networks as a tool for reducing model overfitting.
The experimental reported in this manuscript confirm that the harmonic block outperforms the well established scattering networks using wavelets~\cite{Bruna13,Oyallon18} when limited data is available for training. We provide the PyTorch implementation of the improved harmonic block.
Future work will investigate the effect of window functions that are also often used in Modified DCT as part of the harmonic block, and test its performance in large scale experiments. 

\bibliographystyle{ieeetr}
\bibliography{bibliography}
\end{document}